\documentclass{article}
\usepackage{amsmath,amssymb,amsfonts}
\usepackage{arxiv}
\usepackage{algorithm}
\usepackage{algorithmic}
\usepackage[utf8]{inputenc} 
\usepackage[T1]{fontenc}    
\usepackage{hyperref}       
\usepackage{url}            
\usepackage{booktabs}       
\usepackage{nicefrac}       
\usepackage{microtype}      
\usepackage{lipsum}		
\usepackage[numbers]{natbib}
\usepackage{doi}
\usepackage{graphicx}   
\usepackage{caption}    
\usepackage{cleveref}   
\usepackage{array}
\usepackage{tabularx}
\usepackage{adjustbox}
\usepackage{enumitem}

\newcommand{\figref}[1]{\hyperref[#1]{Figure\ref*{#1}}}

\newcommand{\tabref}[1]{\hyperref[#1]{Table\ref*{#1}}}

\newcommand{\Algref}[1]{\hyperref[#1]{Algorithm\ref*{#1}}}

\title{Effect of a Process Mining based Pre-processing Step in Prediction of the Critical Health Outcomes}

\author{ \href{https://orcid.org/0009-0003-8414-2996}{\includegraphics[scale=0.06]{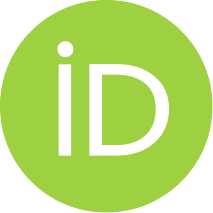}\hspace{1mm}Negin Ashrafi} \\
	Department of Industrial and Systems Engineering\\
	University of Southern California\\
	Los Angeles, CA 90089 \\
	\texttt{ashrafin@usc.edu} \\
	\And
	{\href{https://orcid.org/0009-0007-1387-0995}{\includegraphics[scale=0.06]{orcid.pdf}\hspace{1mm}Armin Abdollahi}} \\
	Department of Electrical Engineering\\
	University of Southern California\\
	Los Angeles, CA 90089 \\
	\texttt{arminabd@usc.edu} \\
	\AND
	\href{https://orcid.org/0000-0003-0307-4075}{\includegraphics[scale=0.06]{orcid.pdf}\hspace{1mm}Greg Placencia} \\
	Department of Industrial and Manufacturing Engineering\\
	California State Polytechnic University-Pomona\\
	Pomona, CA 91768 \\
	\texttt{gvplacencia@cpp.edu} \\
	\And
	\href{https://orcid.org/0009-0003-7159-3245}{\includegraphics[scale=0.06]{orcid.pdf}\hspace{1mm}Maryam Pishgar} \\
	Department of Industrial and Systems Engineering\\
	University of Southern California\\
	Los Angeles, CA 90089 \\
	\texttt{pishgar@usc.edu} \\
}

\begin{document}
\maketitle
\begin{abstract}
Predicting critical health outcomes such as patient mortality and hospital readmission is essential for improving survivability. However, healthcare datasets have many concurrences that create complexities, leading to poor predictions. Consequently, pre-processing the data is crucial to improve its quality. In this study, we use an existing pre-processing algorithm, concatenation, to improve data quality by decreasing the complexity of datasets. Sixteen healthcare datasets were extracted from two databases - MIMIC III and University of Illinois Hospital - converted to the event logs, they were then fed into the concatenation algorithm. The pre-processed event logs were then fed to the Split Miner (SM) algorithm to produce a process model. Process model quality was evaluated before and after concatenation using the following metrics: fitness, precision, F-Measure, and complexity. The pre-processed event logs were also used as inputs to the Decay Replay Mining (DREAM) algorithm to predict critical outcomes. We compared predicted results before and after applying the concatenation algorithm using Area Under the Curve (AUC) and Confidence Intervals (CI). Results indicated that the concatenation algorithm improved the quality of the process models and predictions of the critical health outcomes.
\end{abstract}



\maketitle

\section{Introduction}

A patient’s health records can be understood as a sequence of observations including services performed, diagnoses, or lab measurements \cite{b1}. In the process mining field, each of these observations is called an event, and the sequences of observations are referred to as event logs. Process mining analyzes and optimizes processes. The process mining approach has been used widely in healthcare to enhance healthcare processes \cite{b2} and to predict critical health outcomes \cite{b1,b3,b4,b5}.

Real-life event logs are complex and noisy. Infrequent behaviors and various concurrences generate inefficient and complex process model through process discovery algorithms. Healthcare data exacerbates these issues. For instance, various lab measurements can be taken at the same time causing concurrent relationships among these different lab measurement events. It is therefore critical to pre-process raw event logs to improve their quality, to make the process model more understandable.

A variety of pre-processing algorithms have been used to improve data quality \cite{b6,b7,b26}. The concatenation algorithm is a recent innovation that significantly improves event logs and process model quality for real-life benchmark datasets \cite{b7}. However, the concatenation algorithm has not been tested on real-life healthcare event logs.

Early prediction of patient health outcomes is critical to clinical decision making and effective allocation of medical resources \cite{b27}. For example, Coronary Artery Disease (CAD) causes approximately 610,000 deaths yearly in the United States alone and 17.8 millions deaths worldwide \cite{b8,b25}. Moreover, approximately, 6 million American adults over the age of 20 had heart failure (HF) between 2015 to 2018 \cite{b9,b24, b28}. This has resulted in the Centers for Medicare \& Medicaid Services penalizing hospitals nearly 2.5 billion USD since 2012 for unplanned 30-day readmission rates \cite{b10}. Patient mortality for Paralytic Ileus (PI) can be as high as 40 in ICU settings \cite{b11}. In 2019, diabetes directly caused 1.5 million deaths. More recently, COVID-19, caused by the SARS-CoV-2, heavily impacted morbidity and mortality with major economic consequences \cite{b12}. These highlight the need for accurate predictions models to reduce patient mortality rates and readmission, so as to reduce financial consequences.

Many frameworks have been used to predict the critical health outcomes \cite{b13, b29}. Decay Replay Mining (DREAM) is a recent process mining / deep learning framework. It has significantly improved predicting healthcare outcomes for ICU patients with various diseases \cite{b3,b5} using raw and complex healthcare event logs. However, the effectiveness of pre-processing has not yet been studied using real life healthcare datasets. This work focuses on two issues: 1) Does pre-processing healthcare datasets improve data quality and enhance the process model? 2) Is the concatenation algorithm an effective pre-processor that predicts critical health outcomes such as patient mortality and readmissions? 

The structure of this paper is as follows:  Section 2 provides the preliminaries. Section 3 details the methodology used in this study. Section 4 presents the evaluation, covering Datasets, Setup, Results, and Discussion. Section 5 concludes the findings and directions for future research.

\section{Preliminaries}
The notations used in this section are adopted from \cite{b14}.
\subsection{Process Mining}
Process mining analyzes and optimizes process sequences called event logs. Process mining uses the following steps: process discovery, conformance checking, and process enhancement. The process discovery step is the most important step in process mining. This step uses process discovery algorithms to output a process model, mostly called a Petri Net (PN).

\subsection{Petri Net}
A PN is a mathematical model often used to represent a process. It includes a set of places graphically represented as circles and transitions as rectangles. Transitions correspond to events. The occurrence of a process is modeled by firing transitions. Places can be marked or unmarked. Marked places indicate current process states. A place is marked by putting a dot (token) in the circle corresponding to that event. For a transition to be fired, all its input places must contain a token. If a transition fires, the all tokens are removed from its inputs. Also, each output of the transition receives a token. 

\subsection{Event Logs}
An event $a$ $\epsilon$ $A$ is an instantaneous change of a process’ state where $A$ is the finite set of all possible events. An event $a$ can happen more than once in a provided process. An event instance $E$ is a vector with two attributes at minimum: the name of the associated event $a$ and the corresponding occurrence timestamp, $\tau$. The timestamps $\tau$ of two events cannot be equal. A trace $g$ $\epsilon$ $G$ is a finite and ordered sequence of event instances \cite{b14}. 

\subsection{Decay Replay Mining}
Decay Replay Mining (DREAM) \cite{b15} is a process mining / deep learning-based approach used to predict the next event in the sequence of a process. DREAM uses places from a process model and extends these places with a time decay function $F(\tau)$. The time decay function $F(\tau)$ uses timestamps $\tau$ information as a parameter when replaying event logs $L$ on a process model and produces Timed State Samples (TSS).

\subsection{Timed State Samples}
A TSS will be produced by replaying event logs $L$ on a process model \cite{b15}. A TSS, $S(\tau)$ at time $\tau$ results from concatenating time decay function values $F(\tau)$, token counts $C(\tau)$, and of a marking $M(\tau)$:
\begin{equation}
    S(\tau) =F(\tau) \oplus C(\tau) \oplus M(\tau)
\end{equation}
\subsection{Concatenation Algorithm}
The concatenation algorithm \cite{b7} is a pre-processing algorithm that removes some concurrences and self-loops based on a probability algorithm. This improves the quality of event logs and the resulting process model. 

%

\subsection{Concurrency Relation}
Given any event logs $L$, and any two labels $\ell$, $\ell\prime$ $\epsilon$ $L$, and any two events $e_i$, $e_j$ $\in$ $E$, a concurrent relation exists, denoted by $(e_i  || e_j )$, given the following conditions:
$|\ell  \to \ell \prime| > 0$ and $|\ell \prime \to \ell | > 0 $.

\subsection{Self-loops}
A self-loop exists in the event logs $L$, if $|\ell \to \ell| $ is positive for some event $e$ $\in$ $E$ with $\lambda(e)$ $=$ $\ell$ \cite{b16}.

\subsection{F-Measure}
An F-Measure is an evaluation metric uses to measure the quality of a process model. The F-Measure is a trade-off value between fitness and precision \cite{b17}.
The Fitness of a process model represents the behavior of the event logs $L$. Precision shows the capability of a model to produce the behaviors found only in the event logs $L$. In this work, the alignment-based method is used to measure both fitness and precision per \cite{b14,b21}.

\subsection{Complexity}
Complexity is an evaluation metric that measures the quality of a process model. Complexity shows how easily a process model can be understood. Several metrics measure the complexity of a model \cite{b16}, including Control-Flow Complexity (CFC) \cite{b17}, size of the models \cite{b18}, and structuredness \cite{b18}. CFC shows how many branches are prompted by split gateways in a process model. The size of a process model calculates the numbers of the nodes and arcs in the model. A process model is structured, if, for any split gateway in the process model a corresponding join gateway exits.

\subsection{Split Miner}
Split Miner (SM) is a process discovery algorithm that produce a sound PN from the event logs $L$. It produces a process model with less complexity and higher F-Measure value than other process discovery algorithms. The SM algorithm includes the following steps: First, generate a directly-follow graph and identify short loops. Second, discover the existing concurrency relation between events. Third, utilize filtering by introducing 2 threshold values, to control the filtering process (frequency threshold, $\epsilon$), the other, $\eta$, to control concurrency relations. Moreover, the algorithm derives choice and concurrency relations by adding split gateways. In the end, the join gateways are discovered. The output of these steps is a BPMN, and it can be converted to a PN through using ProMs’ BPMN Miner package \cite{b19}. This algorithm is publicly available as a java application \cite{b19}.

\section{Methodology}
This section focuses on a proposed method for predicting several health outcomes with and without applying a pre-processing step on event logs $L$. First, feature selection is described. Next, conversion of Electronic Health Records (EHR) to event logs $L$ is explained followed by the pre-processing step. Finally, the DREAM algorithm is introduced to predict health outcomes.

\begin{figure}[htbp]
    \centering
    \includegraphics[width=8cm, height=7cm]{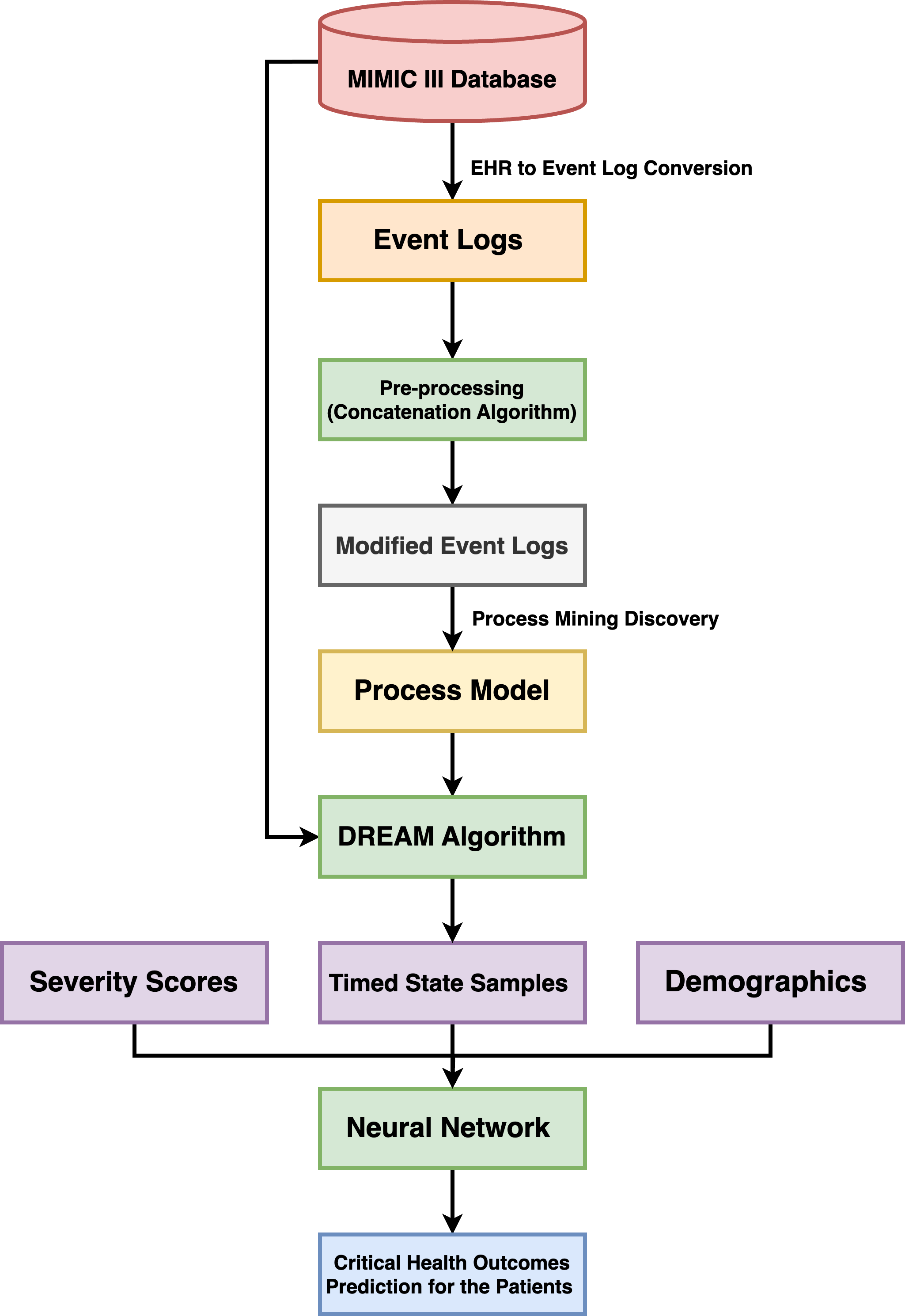}
    \caption{Overview of the Proposed Model}
    \label{fig:1}
\end{figure}

The intuition behind this methodology is that applying an appropriate pre-processing step on complex healthcare event logs $L$ decreases complexity of a process model. Hence, it predicts critical health outcomes more accurately. Since healthcare event logs $L$ contain many concurrences and self loops, applying a concatenation algorithm to concurrent events makes the process model simpler and generates more accurate predictions. An overview of the proposed framework is shown in \figref{fig:1}.

\subsection{Feature Selection}
MIMIC III \cite{b20} and University of Illinois Hospital (UIH) databases\cite{b13} were used for predictions. Due to the nature of our approach (process mining / deep learning approach) every patient must have a medical history available at the time of prediction. \tabref{table_1} shows the features extracted from MIMIC III and UIH databases by disease category.

\renewcommand{\arraystretch}{1.5} 

\begin{table*}[ht]
\caption{Features Extracted from MIMIC III and UIH Databases}
\label{table_1}
\centering
\scriptsize 
\begin{adjustbox}{width=1\textwidth}
\begin{tabularx}{\textwidth}{|>{\centering\arraybackslash}X|>{\centering\arraybackslash}X|>{\centering\arraybackslash}X|>{\centering\arraybackslash}X|>{\centering\arraybackslash}X|>{\centering\arraybackslash}X|>{\centering\arraybackslash}X|>{\centering\arraybackslash}X|}
\hline
\multicolumn{8}{|c|}{\textbf{Data Type (Part 1)}} \\
\hline
\textbf{Type of Disease} & \textbf{Admission} & \textbf{Insurance} & \textbf{Lab Related} & \textbf{Demographics} & \textbf{Elixhauser Comorbidities} & \textbf{CPT and ICD-9-CM codes} & \textbf{Discharge} \\
\hline
\textbf{PI} & x & x & x & x & NA & NA & x \\
\hline
\textbf{HF} & x & x & x & x & x & x & x \\
\hline
\textbf{COVID} & & & x & x & & & \\
\hline
\textbf{Diabetes} & x & x & x & x & x & x & NA \\
\hline
\textbf{CAD} & x & x & x & x & x & NA & NA \\ 
\hline
\end{tabularx}
\end{adjustbox}

\vspace{0.5cm}

\begin{adjustbox}{width=1\textwidth}
\begin{tabularx}{\textwidth}{|>{\centering\arraybackslash}X|>{\centering\arraybackslash}X|>{\centering\arraybackslash}X|>{\centering\arraybackslash}X|>{\centering\arraybackslash}X|>{\centering\arraybackslash}X|>{\centering\arraybackslash}X|>{\centering\arraybackslash}X|}
\hline
\multicolumn{8}{|c|}{\textbf{Data Type (Part 2)}} \\
\hline
\textbf{Type of Disease} & \textbf{Severity scores} & \textbf{Care unit in and out from ICU} & \textbf{Location} & \textbf{Encounter-based} & \textbf{Report-based} & \textbf{Vitals signs} & \textbf{ICU-based} \\
\hline
\textbf{PI} & NA & x & & & & & \\
\hline
\textbf{HF} & x & & & & & & \\
\hline
\textbf{COVID} & & & x & x & x & x & x \\
\hline
\textbf{Diabetes} & x & NA & & & & & \\
\hline
\textbf{CAD} & NA & NA & & & & & \\ 
\hline
\end{tabularx}
\end{adjustbox}
\end{table*}

\subsection{Conversion of EHR to the Event Logs}
Event logs $L$ can be understood as a sequence of events with timestamps $\tau$ associated to events that have occurred. Events represent activities performed on patients such as admission, diagnosis, lab measurements, etc., also known as patients careflows. Each unique event has a corresponding timestamp $\tau$ when it occurred. In this paper, the timestamp $\tau$ of comorbidity and artificial events correspond to the discharge time of the corresponding hospital admission. Since events in a trace $g$ occur sequentially, multiple comorbidities or artificial events with the same timestamps $\tau$ are delayed by multiples of 1 ms to retain their order. \tabref{table_2} summarizes the conversion of the EHR to the event logs $L$.
\renewcommand{\arraystretch}{1} 
\begin{table*}[ht]
\caption{Conversion of EHR from Databases to Event Logs}
\label{table_2}
\centering
\small
\begin{adjustbox}{width=0.85\textwidth,center}
\begin{tabular}{|>{\centering\arraybackslash}p{0.4\textwidth}|>{\centering\arraybackslash}p{0.4\textwidth}|}
\hline
\textbf{MIMIC III and UIH Tables} & \textbf{Events} \\
\hline
ADMISSIONS, PATIENTS & 
\begin{tabular}[c]{@{}c@{}}Admission type, Demographics,\\ patient's exit type, discharge, insurance type\end{tabular} \\
\hline
D\_LABITEMS, LABEVENTS, ADMISSIONS & Lab events \\
\hline
ICUSTAYS, CALLOUT & Careunit type \\
\hline
LABEVENTS & Abnormal flagged events \\
\hline
Admissions, Diagnoses\_ICD & Elixhauser Comorbidity Score events \\
\hline
Diagnose\_ICD, Procedures\_ICD, CPTevents & 30 Artificial event abstractions \\
\hline
\end{tabular}
\end{adjustbox}
\end{table*}
\subsection{Pre-processing Step}

In \Algref{alg:cat}, the resultant event logs ($L$) from the previous step were then used as an input to the concatenation algorithm. Assume that events $e_i$, $e_j$, $e_k$, and $e_l \in L$, if $(e_i \rightarrow e_j)$ and $(e_j \rightarrow e_i)$; also $(e_k \rightarrow e_l)$ and $(e_l \rightarrow e_k)$, if there is a concurrent relations between $e_i$, $e_j$, $e_i || e_j$, and between $e_k$, $e_l$, $e_k || e_l$.

\begin{algorithm}
\caption{Event Log Concatenation Algorithm}
\label{alg:cat}
\centering
\begin{algorithmic}[1]
\REQUIRE Event logs $L$, threshold $p^*$
\ENSURE Modified event logs $L'$

\STATE Initialize an empty list of concurrent pairs
\FORALL{events $e_i, e_j \in L$}
    \IF{$(e_i \rightarrow e_j) \land (e_j \rightarrow e_i)$}
        \STATE Add $(e_i||e_j)$ to the list of concurrent pairs
    \ENDIF
\ENDFOR

\STATE Initialize an empty list of valid pairs
\FORALL{pairs $(e_i||e_j)$ in the list of concurrent pairs}
    \IF{$P(e_i \rightarrow e_j) > p^*$ and $P(e_j \rightarrow e_i) > p^*$}
        \STATE $P(e_i||e_j) = P(e_i \rightarrow e_j) + P(e_j \rightarrow e_i)$
        \STATE Add $(e_i||e_j, P(e_i||e_j))$ to the list of valid pairs
    \ENDIF
\ENDFOR

\STATE Sort the list of valid pairs in descending order of $P(e_i||e_j)$

\FORALL{pairs $(e_i||e_j)$ in the sorted list of valid pairs}
    \STATE Remove $e_i$ and replace $e_j$ with $e_i \oplus e_j$
\ENDFOR

\FORALL{events $e_i \in L$ not concatenated}
    \IF{$e_i \in e_i \oplus e_j$}
        \STATE Remove $e_i$ and replace with $e_i \oplus e_j$
    \ENDIF
\ENDFOR

\FORALL{events $e_i \in L$}
    \IF{$e_i \rightarrow e_i$}
        \STATE Remove $e_i$
    \ENDIF
\ENDFOR

\STATE Feed the modified event logs $L'$ to the SM algorithm for process modeling
\STATE Use the resultant process model as input to the DREAM algorithm for prediction

\end{algorithmic}
\end{algorithm}

We then calculate $P(e_i \rightarrow e_j)$ and $P(e_j \rightarrow e_i)$; and $P(e_k \rightarrow e_l)$ and $P(e_l \rightarrow e_k)$. Moreover, threshold value $p^*$ is introduced, if $P(e_i \rightarrow e_j)>p^*$ and $P(e_j \rightarrow e_i)>p^*$, the combination is chosen. Similarly, if $P(e_k \rightarrow e_l)>p^*$ and $P(e_l \rightarrow e_k)>p^*$, the combination is chosen for the next step. Additionally, the probabilities of the selected combinations were added and sorted in the descending order, $P(e_i||e_j)=P(e_i \rightarrow e_j) + P(e_j \rightarrow e_i)$ also $P(e_{k}||e_{l})=P(e_{k} \rightarrow e_{l})+P(e_{l } \rightarrow e_{k})$. Furthermore, if $P(e_i||{e_j})= P(e_{k}||e_{l})$ a re-position step was applied. After the order of combinations for concatenation is finalized, the concatenation algorithm selects the ordered combinations one by one for the concatenation. Assume that  $(e_i,$ $e_j)$ is chosen for concatenation, event $e_i$ is removed and replaced with the concatenated event, $e_i \oplus e_j$. This step is repeated until all selected combinations are concatenated in the event logs $L$. After concatenating all combinations in the event logs $L$ based on descending order, some events will remain that are not concatenated. Each non concatenated event will be examined and checked for event, $e_i \in$ $e_i \oplus e_j$. In that case, $e_i$ will be removed and replaced with $e_i \oplus e_j$. If $e_i$ $  \notin $ $  e_i \oplus e_j$, while the final trace remains unchanged. Finally, if $e_i \rightarrow e_i$ exist, it is removed. The modified event logs $L$ was then fed to the SM algorithm to produce a process model. The resultant process model was input to the DREAM algorithm for prediction as detailed in the next subsection.

\subsection{DREAM algorithm}
The algorithm has three different phases \cite{b15}. The first phase is to create a PN model from an event log $L$ and assign a decay function to each PN location $f_p(\tau)$. The discovery log is then replayed, and feature arrays comprising decay function response values, token movement counts, and resource use are extracted. Lastly, we train a Neural Network (NN) to predict the next event, using these feature arrays as training data. To simulate data values that decrease over time, one can use decay functions. Financial domains, physical systems, and population trend modeling are among the common applications for these functions. We assign a linear decay function $f_p(\tau)$ to every place in the PN produced by SM on an event log $L$.

\begin{algorithm}
\caption{PN Discovery and Decay Function Initialization}
\label{alg:pn-discovery-decay}
\begin{algorithmic}[1]
\REQUIRE Event log $L$ with traces $\{g_1, g_2, \ldots, g_n\}$
\REQUIRE Decay function parameters $\alpha, \beta$
\ENSURE Discovered PN with initialized decay functions

\STATE \textbf{Step 1: PN Model Discovery}
\STATE Use SM to discover a PN model from event log $L$
\FOR{each place $p$ in the PN model}
    \STATE Associate place $p$ with a linear decay function $f_p(\tau) = \max(\beta - \alpha \cdot \Delta_p, 0)$
\ENDFOR

\STATE \textbf{Step 2: Decay Function Enhancement}
\FOR{each place $p$ in the PN model}
    \STATE Initialize the decay function $f_p(\tau)$ to $0$ by setting $\Delta_p = \infty$
\ENDFOR

\STATE \textbf{Step 3: Calculate Decay Rates}
\FOR{each place $p$ in the PN model}
    \STATE Calculate the maximum trace duration $\Delta_{\text{max}}(L) = \max_{1 \leq i \leq |L|} (\nu_{d_{ts}}(L_{i,\gamma(L_i)}) - \nu_{d_{ts}}(L_{i,1}))$
    \STATE Define $v_p(g)$ as the number of tokens entering place $p$ during replay of trace $g$
    \IF{$\max_{g \in L} v_p(g) \leq 1$}
        \STATE Set $\alpha_p = \frac{\beta}{\Delta_{\text{max}}(L)}$
    \ELSE
        \STATE Set $\alpha_p = \frac{\beta}{\text{mean}_{g \in L} \delta_p(g)}$
    \ENDIF
\ENDFOR

\STATE \textbf{Step 4: Update Decay Functions with Calculated Rates}
\FOR{each place $p$ in the PN model}
    \STATE Update the decay function $f_p(\tau) = \max(\beta - \alpha_p \cdot (\tau - \tau_p), 0)$
\ENDFOR

\STATE Return the discovered PN model with initialized decay functions

\end{algorithmic}
\end{algorithm}

By using $\Delta_p = \tau - \tau_p$, we can express the time difference between the current time, $\tau$, and the most recent time a token entered place p, $\tau_p$. With the help of the two decay function parameters, $\alpha$ and $\beta$, we can adjust the significance level. Based on the reactivation durations of a place p, $\alpha$ should ideally be set so that the slope of $f_p(\tau)$ covers the entire range from $\beta$ to 0. In other words, the slope should not be excessively steep for a small $\Delta_p$ such that $f_p(\tau) = 0$ or too flat for a large $\Delta_p$ such that $f_p(\tau) = B$. This suggests varying $\alpha$ values with each place in the PN model.
In the next step of \Algref{alg:pn-discovery-decay} we try to estimate these $\alpha_p$ values. Using the event log $L$ and the corresponding PN found by the SM on $L$, we estimate $\alpha_p$. There are a limited number of event instances in each trace $g$ in $L$. The $j^{\text{th}}$ event instance of an event log L's $i^{\text{th}}$ trace is referred to by $L_{i,j}$. $\lambda(L_i)$ expresses the number of event instances of the $i^{\text{th}}$ trace of the event log $L$. The value of the attribute d for the event instance E is $\nu_d (E)$. When an attribute d is absent from an event instance E, the result is $\nu_d (E) = 0$, an empty set. We use $d_{ts}$ to represent the attribute timestamp. $\Delta_{max}(L)$, the Maximum observed trace duration in an event log $L$, is defined in the algorithm and based on the value of $\max_{g \in L} v_p(g)$ the value of $\alpha_p$ is determined. If the value is less than one then $\alpha_p$ will be set to a value that ensures that before the final event instance of a corresponding trace $g$ in $L$ occurs, the response of $f_p(\tau)$ will never equal 0. By doing this, we ensure that, until the end of a trace, we will have information about the occurrence of a particular event in the decay function's response. However, when the value of $\max_{g \in L} v_p(g)$ is larger than one we calculate the average time it takes for a place p to reactivate. With this knowledge, we adjust the decay rate so that, for the average time between reactivations, $f_p(\tau)$ yields a level of recent token movement importance.

\begin{algorithm}
\caption{Event Log Replay and Feature Extraction}
\label{alg:replay-feature-extraction}
\begin{algorithmic}[1]
\REQUIRE Event log $L$, PN with places $P$, decay functions $f_p(\tau)$ for each place $p \in P$
\ENSURE Set of timed state samples $S$

\STATE Initialize counting vector $C(\tau)$ to zero
\FOR{each trace $g \in L$}
    \STATE Reset $F(\tau)$, $C(\tau)$, and marking $M(\tau)$ to initial states
    \FOR{each event $e_i$ in trace $g$}
        \STATE Calculate decay function responses $F(\tau) = \{f_p(\tau)\ |\ p \in P\}$
        \STATE Update $C(\tau)$ for token movements
        \STATE Update PN marking $M(\tau)$ based on transition fired by $e_i$
        \STATE Construct timed state sample $S(\tau) = F(\tau) \oplus C(\tau) \oplus M(\tau)$
        \STATE Add $S(\tau)$ to set of timed state samples $S$
    \ENDFOR
\ENDFOR

\STATE Return set of timed state samples $S$

\end{algorithmic}
\end{algorithm}

As a result, the slope will not be overly flat or steep.
In \Algref{alg:replay-feature-extraction} $F(\tau)$ is a vector of decay functions for all places. $C(\tau)$ shows the number of tokens
that have entered each place from time 0 to $\tau$. When a token enters place p and time $\tau$ the corresponding value in $C(\tau)$ vector is updated. Here $M(\tau)$ is a vector representing the marking of a PN. So for each trace $g$ and events in traces, the value of the aforementioned vectors are updated, and then the timed state sample $S(\tau)$ is constructed by concatenating the vectors. Consequently, a timed state sample, $S(\tau)$ provides a time-dependent description of a PN process state. It includes data regarding time-based token movements, such as token counts per location (loop information), the current PN state using the marking, and the last time a token entered a place in relation to the current time. In the last step of the algorithm, the $S(\tau)$ is input to the different NN with different architecture for predicting the next event.

\subsection{Prediction}
DREAM was used to predict the critical health outcomes for both pre-processed and non-processed event logs $L$. DREAM replays the event logs $L$ on a process model and produces time information related to the variables which are called TSS. The TSS, demographic information, and severity scores were fed into the dense NN for prediction. The NN architecture differs by disease but remaine the same for both pre-processed and unprocessed event logs $L$.

\subsubsection{CAD disease} 
\begin{figure}[H]
  \includegraphics[width=8.25cm, height=3cm] 
  {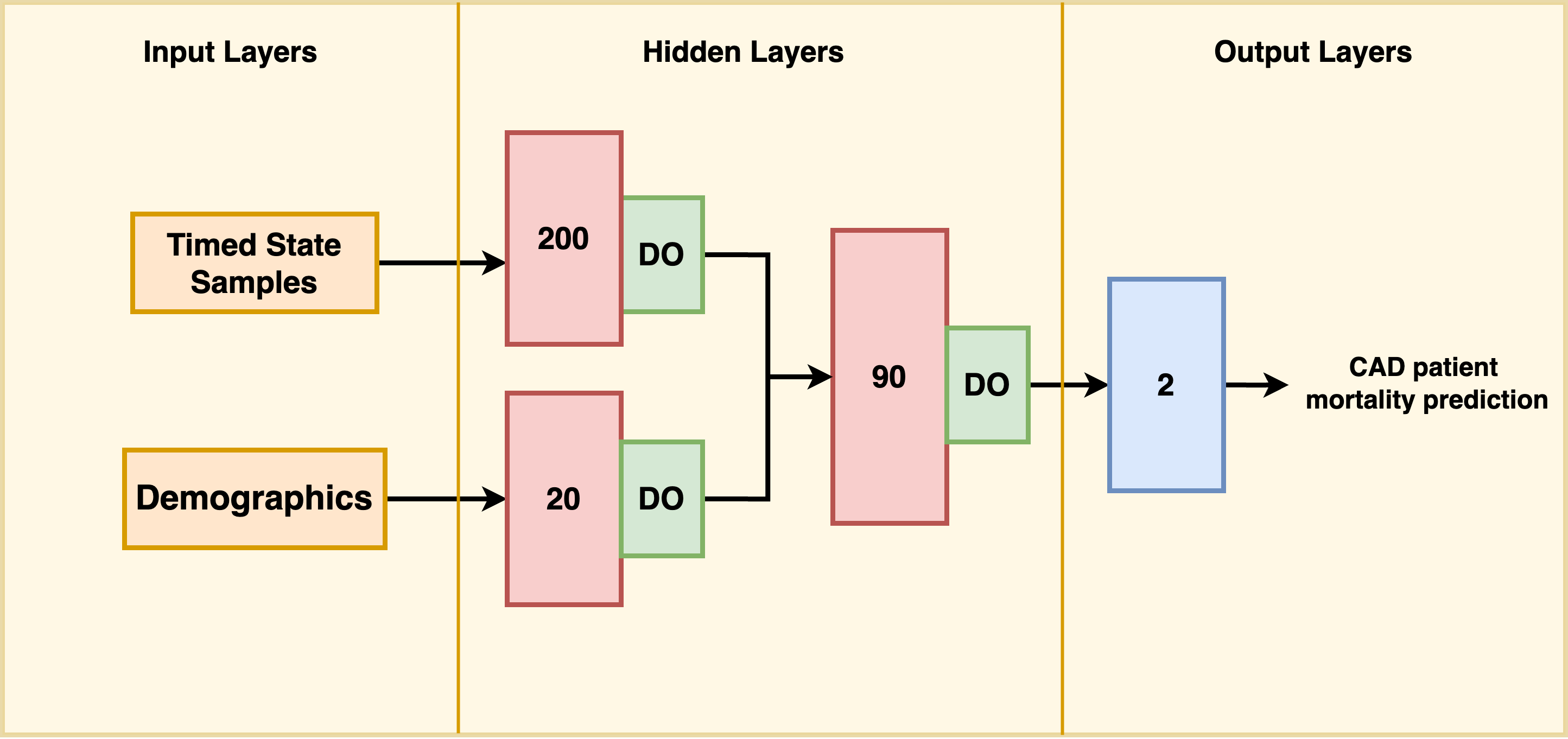}
\centering
\caption{Architecture of the NN to predict CAD related outcome}
\label{fig:2}       
\end{figure}

TSS are fed into a unique branch of one hidden layer with $200$ neurons as shown in \figref{fig:2}. The Dropout (DO) rate after the first layer was regularized at DO = 0.2. Demographic information was fed into a single hidden layer with $20$ neurons and the same dropout rate. These were concatenated into a second layer with $90$ neurons and the same dropout rate. The critical outcome which is predicted in this case is the mortality of ICU patients with CAD.

\subsubsection{HF disease} 
\begin{figure}[H]
  \includegraphics[width=8.25cm, height=4cm] 
  {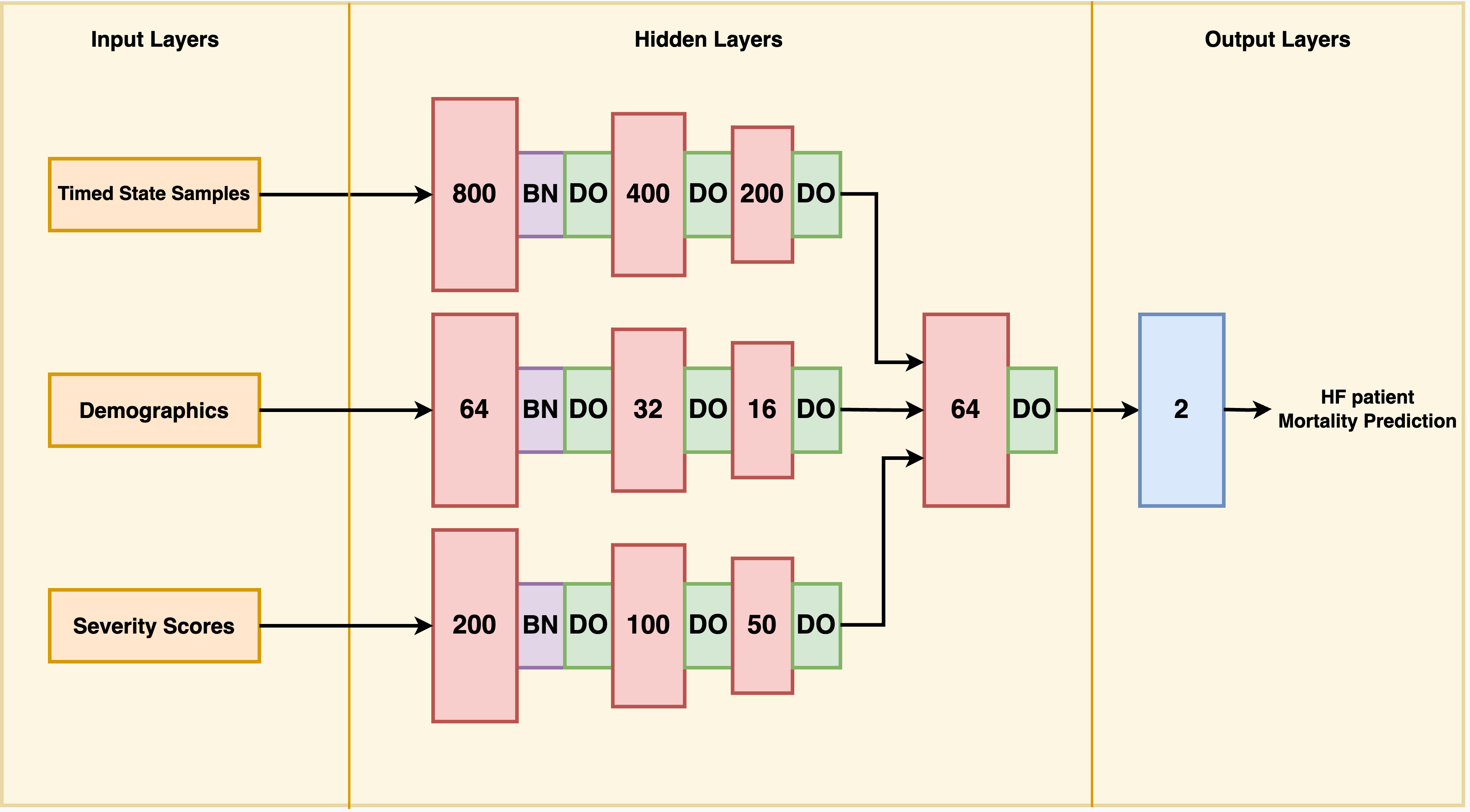}
  \centering
\caption{Architecture of the NN to predict HF related outcome}
\label{fig:3}       
\end{figure}

TSS, demographics information, and severity scores were fed separately to three branches each with three hidden layers (\figref{fig:3}). A batch normalization layer was added after the first hidden layer for each branch with a dropout rate of 0.2. The output layer included a softmax activation function to predict unplanned 30-day readmission of ICU HF patients. The critical outcome predicted in this case was unplanned 30-day readmission of ICU patients with HF.

\subsubsection{Diabetes} 
\begin{figure}[H]
  \includegraphics[width=8.25cm, height=4cm] 
  {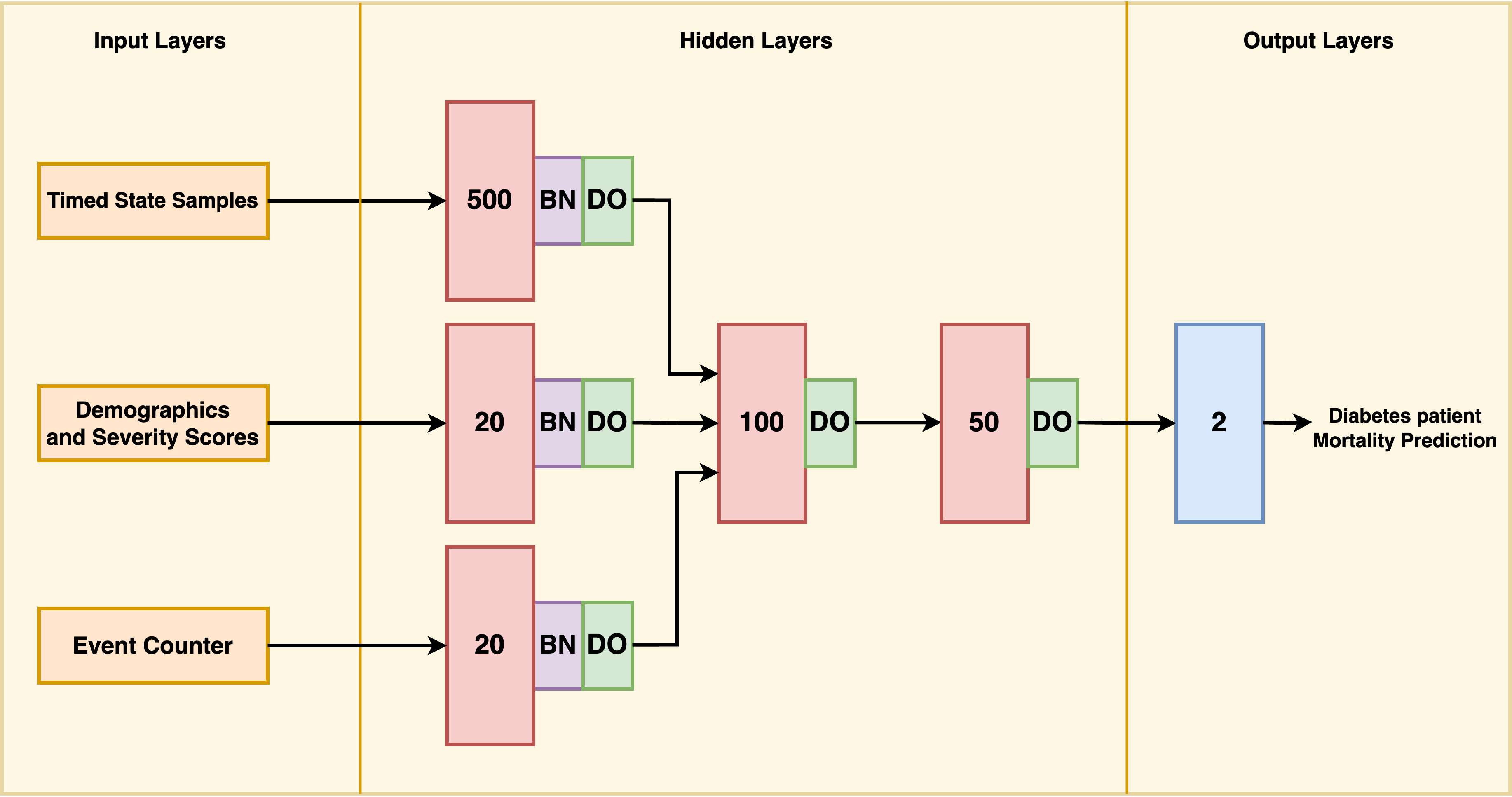}
  \centering
\caption{Architecture of the NN to predict Diabetes related outcome}
\label{fig:4}       
\end{figure}

Each input layer is fed to an individual hidden layer branch before a hidden layer concatenates the branches and feeds it to two further layers. All hidden layers use a Rectified Linear Unit activation function (ReLU). The branched first hidden layers also have a batch normalization layer. Dropout layers with a rate of 0.4 for regularization are used. The output layer consists of a softmax activation function to output the patient’s probability of in-hospital mortality for diabetes patients. This architecture is shown in \figref{fig:4}.

\subsubsection{COVID- 19} 
\begin{figure}[H]
  \includegraphics[width=8.25cm, height=4cm] 
  {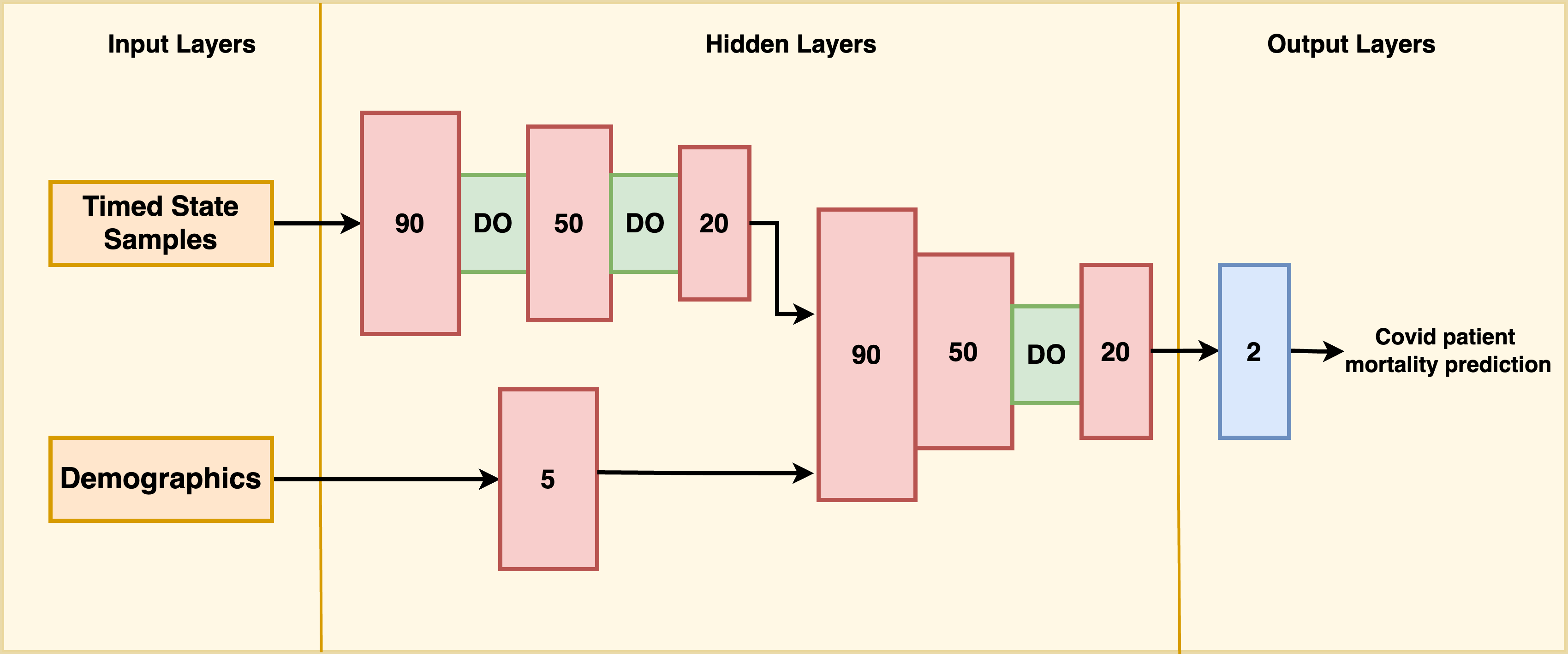}
  \centering
\caption{Architecture of the NN to predict COVID-19 related outcome}
\label{fig:5}       
\end{figure}

TSS, demographics information and comorbidities were fed separately to two branches where the first branch contained three hidden layers with 90, 50 and 20 neurons respectively (\figref{fig:5}). The first and second hidden layers each had a dropout layer with a rate of 0.2. The second branch also contained one hidden layer with 5 neurons. The two branches were concatenated to a branch with three hidden layers, containing 90, 50, and 20 neurons respectively. The dropout layer after the second concatenated hidden layer had a rate of 0.3. The output layer used a softmax activation function to predict mortality of COVID- 19 patients every 6-hour during the first 72 hours after hospital admission.

\subsubsection{PI} 
\begin{figure}[H]
  \includegraphics[width=8.25cm, height=4cm] 
  {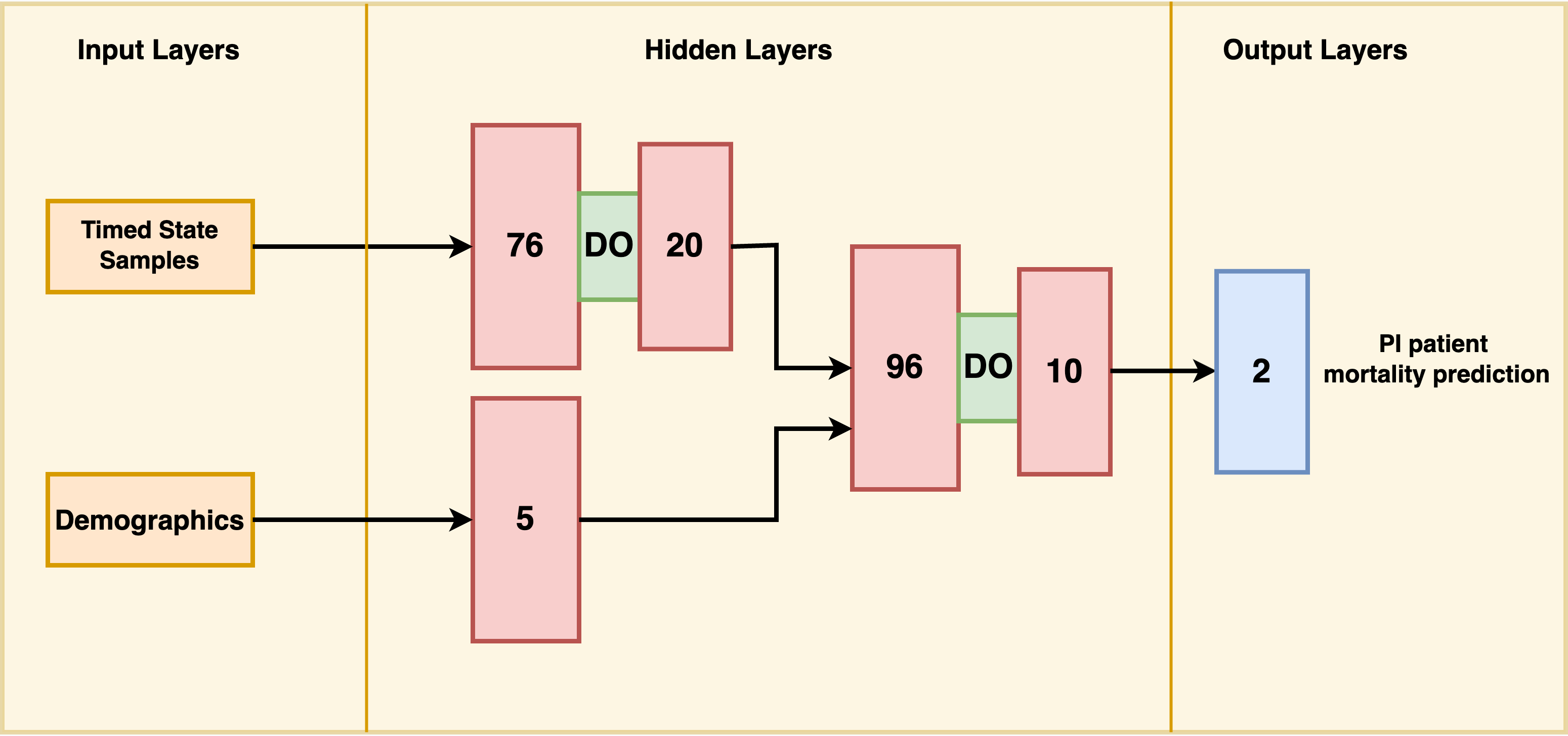}
  \centering
\caption{Architecture of the NN to predict PI related outcome}
\label{fig:6}       
\end{figure}

In \figref{fig:6}, TSS were fed into two hidden layers. The first had 76 neurons and a subsequent dropout rate of DO = 0.5. The second hidden layer contained 20 neurons. The demographic information was fed into a single hidden layer of 5 neurons. Both layers were concatenated into two further layers with 96 and 10 neurons respectively. Between these two layers was a layer with a dropout rate of DO = 0.5. The outcome predicted in this case was the mortality of ICU patients diagnosed with PI 24 hours after being admitted to the ICU.

\section{Evaluation}
This section discusses experimental evaluation of the model to predict the critical health outcomes. Subsection one describes the datasets. Subsection two describes the modeling set up. Subsection three highlights the results.

\subsection{Datasets}
Data was obtained from MIMIC III (Medical Information Mart for Intensive Care) \cite{b20}, a large database containing information relating to patients admitted to Beth Israel Deaconess Medical Center (BIDMC). This data includes vital signs, medications, laboratory measurements, observations and notes charted by care providers, fluid balance, procedure codes, diagnostic codes, imaging reports, hospital length of stay, survival data, and more \cite{b20}. Data was also obtained from UIH, a tertiary, teaching hospital in Chicago. This study was approved by The University of Illinois at Chicago (UIC) Institutional Review Board \cite{b13}. A summary of the datasets for each disease category is shown in \tabref{table_3}.

\begin{table}[H]
\caption{Summary of datasets by disease category}
\label{table_3}
\centering
\renewcommand{\arraystretch}{1.5} 
\begin{adjustbox}{width=0.6\textwidth}
\begin{tabular}{|l|c|c|c|c|}
\hline
Disease & Total Numbers of Patients & Train & Test & Validation\\
\hline
HF & 3,411 & 2,422 & 555 & 434\\
\hline
COVID-19 & 508 & 303 & 104 & 101\\
\hline
PI & 1,153 & 681 & 336 & 136\\
\hline
CAD & 2,176 & 1,202 & 674 & 300\\
\hline
Diabetes & 2,435 & 1,552 & 609 & 274\\
\hline
\end{tabular}
\end{adjustbox}
\end{table}

\subsection{Setup}
RapidProM was used to run experiments \cite{b21} extending RapidMiner with process mining analysis capabilities. A concatenation algorithm's threshold value of $p^*$ was used to select the combinations of concurrent events for further concatenation steps.

The optimal value of $p^*$ was based on the highest F-Measure value. The hyper-parameter optimization was conducted on 16 datasets using steps of 0.1. Of these, 13, found an optimal value of $p^* = 0.7$.

SM takes two threshold values: $\eta$ and $\epsilon$. These values were optimized based on the highest F-Measure values using steps of 0.1. Thresholds selected were $\eta = 0.4$ and $\epsilon = 0.1$.

Train and validation sets are required to discover a process model and train the NN. The validation set is used to select the best model. A test set is then used to evaluate model performance. A summary of the set up for training the NN for each disease category is shown in \tabref{table_4}.

\begin{table}[ht]
\centering
\caption{Summary of setup for training NN}
\label{table_4}
\renewcommand{\arraystretch}{1.5} 
\begin{adjustbox}{width=0.6\textwidth}
\begin{tabular}{|l|c|c|c|c|}
\hline
Disease & Activation Function & Epoch & Batch Size & Optimizer\\
\hline
HF & ReLU & 100 & 12 & Adam\\
\hline
COVID-19 & ReLU & 350 & 10 & Adam\\
\hline
PI & ReLU & 350 & 50 & RMSprop\\
\hline
CAD & ReLU & 300 & 56 & RMSprop\\
\hline
Diabetes & ReLU & 200 & 256 & Adam\\
\hline
\end{tabular}
\end{adjustbox}
\end{table}

We measured the quality of the process models with and without pre-processing using common metrics of fitness, precision, and F-Measure as proxies for accuracy, size, CFC, and structuredness, which indicate complexity. 

The metric used to evaluate the model prediction is the AUC score which is equal to the probability that a classifier ranks a randomly chosen positive instance higher than a randomly chosen negative one. AUC is a better classification estimate than other common classification performance metrics \cite{b22}. Higher AUC scores indicate that a model is better at distinguishing between discharged patients and those who die. Furthermore, the 95\% confidence interval (CI) for the AUC score is calculated using DeLong's method \cite{b23}.

\subsection{Results}

In our first experiments without pre-processing, we used the optimal threshold values found for the SM algorithm, plus raw event logs to discover process models. The model was fed to the DREAM algorithm to produce a TSS. The TSS, severity score and demographics were then fed to a NN for prediction. 

For our second experiments, with pre-processing we applied the concatenation algorithm to raw event logs $L$ using optimal thresholds values for the algorithm to produce modified event logs $L$. The modified event logs $L$ were then fed to the SM algorithm with the aforementioned optimal threshold values to discover a process model, which was then fed to the DREAM algorithm to produce the TSS. The TSS, severity score and demographics were then fed to a NN for prediction. 

Results of the evaluation are summarized in \tabref{table_5a}. The F-Measure is calculated for different types of disease before and after the concatenation algorithm. In \tabref{table_5b}, AUC and CI are calculated and in \tabref{table_5c} complexity is measured by evaluating the size of the models, sructuredness and CFC. 
We ran the Wilcoxon Test to determine whether improvements observed for evaluation metrics applied after concatenation were statistically significant or not ($p=0.05$). Of 6 pre-processed datasets, 9 showed statistically significant differences in AUC scores. All 16 showed statistically significant differences in F-Measures and complexity metrics.  
\begin{table}[ht]
\centering
\captionsetup{justification=centering}
\caption{Results in terms of F-Measure on event logs, and the Wilcoxon test results on improved cases to statistically identify significance level}
\label{table_5a}
\renewcommand{\arraystretch}{1.5} 
\begin{adjustbox}{width=0.48\textwidth}
\begin{tabular}{|l|c|c|c|}
\hline
&\multicolumn{3}{c|}{Evaluation Metrics}\\
Type of Disease&
\multicolumn{3}{c|}{F-Measure}\\
\hline
&before concat&
after concat&
p-value\\
\hline
HF & 0.821& 0.830& 0.048\\
\hline
PI & 0.842& 0.861 & 0.039\\
\hline
COVID 6hr & 0.844& 0.871 & 0.037\\
\hline
COVID 12hr & 0.844& 0.871& 0.037\\
\hline
COVID 18hr & 0.844& 0.871& 0.037\\
\hline
COVID 24hr & 0.844& 0.871& 0.037\\
\hline
COVID 30hr & 0.844& 0.871& 0.037\\
\hline
COVID 36hr & 0.844& 0.871& 0.037\\
\hline
COVID 42hr & 0.844& 0.871& 0.037\\
\hline
COVID 48hr & 0.844& 0.871& 0.037\\
\hline
COVID 54hr & 0.844& 0.871& 0.037\\
\hline
COVID 60hr & 0.844& 0.871& 0.037\\
\hline
COVID 66hr & 0.844& 0.871& 0.037\\
\hline
COVID 72hr & 0.844& 0.871& 0.037\\
\hline
Diabetes & 0.791& 0.792& 0.055\\
\hline
CAD & 0.832& 0.854& 0.045\\ 
\hline
\end{tabular}
\end{adjustbox}
\end{table}

\subsection{Discussion}
To reduce concurrences and self-loops of the complex healthcare data, we applied a concatenation algorithm as a pre-processing step to improve data quality to improve the trustworthiness of prediction models for clinicians. We observed significant statistical improvements in AUC values, F-Measure and complexity metrics.

This was especially evident for the CAD and COVID-19 datasets which were more complex compared to other datasets. In those cases the predicted results improved by significant statistical amount of 2\% for the AUC metric from our experiments.


One of the main advantages of a process mining approach is that it discovers a detailed process model which allows clinicians to comprehensively visualize the entire process patients (careflow) go through in a medical system. This is precisely why process mining has been preferred over other traditional machine learning methods. Process mining also accurately models temporal time-related information related to variables and utilizes them effectively as an input to the neural networks when other traditional techniques cannot. Lastly, process mining leverages the extensive and detailed medical history of patients from prior hospital admissions.

A limitation of the proposed approach is that it requires patients' medical history that small clinics and hospitals might not have. It also excludes patients with no prior hospital admissions like new patients and patients who are not admitted before the current hospital visit. Moreover, hospitals tend not to share patient data across outside hospital networks. Thus, the proposed approach best matches large hospital networks. In addition, the training, validation, and test sets all came from the MIMIC-III and UIH datasets. Therefore, using an independent dataset from a different health center or hospital would better test the performance of the model for future research.

\begin{table*}[ht]
\centering
\captionsetup{justification=centering}
\caption{Results of the proposed model in terms of AUC and CI on event logs, and the Wilcoxon test results on the improved cases to statistically identify the level of significance}
\label{table_5b}
\renewcommand{\arraystretch}{1.55} 
\begin{adjustbox}{width=0.6\textwidth}
\begin{tabular}{|l|c|c|c|c|c|}
\hline
&\multicolumn{5}{c|}{Evaluation Metrics}\\
\hline
&\multicolumn{3}{c|}{AUC}&\multicolumn{2}{c|}{CI}\\
\hline
Type of Disease&before concat&after concat&p-value&before concat&after concat\\
\hline
HF & 0.930& 0.947& 0.045 & [0.898- 0.960]& [0.910- 0.971]\\
\hline
PI & 0.810& 0.823& 0.047 &[0.768- 0.840]& [0.788- 0.850]\\
\hline
COVID 6hr & 0.776& 0.779& 0.055 & [0.678- 0.876]& [0.701- 0.790]\\
\hline
COVID 12hr & 0.782& 0.79& 0.043 & [0.685- 0.880]&[0.708- 0.850]\\
\hline
COVID 18hr & 0.806& 0.802& NA & [0.719- 0.901]& [0.788- 0.850]\\
\hline
COVID 24hr & 0.799& 0.791& NA & [0.698- 0.890]& [0.718- 0.870]\\
\hline
COVID 30hr & 0.814& 0.841 & 0.034 & [0.718- 0.910]& [0.786- 0.855]\\
\hline
COVID 36hr & 0.814& 0.799& NA & [0.718- 0.900]& [0.778- 0.890]\\
\hline
COVID 42hr & 0.817& 0.817& NA & [0.701- 0.870]& [0.789- 0.860]\\
\hline
COVID 48hr & 0.806& 0.82& 0.042 & [0.738- 0.890]& [0.784- 0.880]\\
\hline
COVID 54hr & 0.853& 0.869& 0.039 & [0.758- 0.910]& [0.788- 0.890]\\
\hline
COVID 60hr & 0.843& 0.842& NA & [0.768- 0.870]& [0.748- 0.890]\\
\hline
COVID 66hr & 0.875& 0.881& 0.046 & [0.778- 0.940]& [0.798- 0.950]\\
\hline
COVID 72hr & 0.900& 0.915& 0.045 & [0.868- 1.00]& [0.870- 0.990]\\
\hline
Diabetes & 0.873& 0.861& NA & [0.851- 0.940]& [0.788- 0.890]\\
\hline
CAD & 0.871& 0.890& 0.044 & [0.831- 0.913]& [0.870- 0.923]\\
\hline
\end{tabular}
\end{adjustbox}
\end{table*}

\begin{table*}[ht]
\centering
\captionsetup{justification=centering}
\caption{Results of in terms of complexity metrics on event logs, and the Wilcoxon test results on the improved cases to statistically identify the level of significance}
\label{table_5c}
\renewcommand{\arraystretch}{1.5} 
\begin{adjustbox}{width=0.9\textwidth}
\begin{tabular}{|l|c|c|c|c|c|c|c|c|c|c|}
\hline
&\multicolumn{9}{c|}{Evaluation Metrics}\\
\hline
&\multicolumn{3}{c|}{Size}&
\multicolumn{3}{c|}{Struct.}&
\multicolumn{3}{c|}{CFC}\\
Type of Disease&
before concat&
after concat&
p-value&
before concat&
after concat&
p-value&
after concat&
before concat &
p-value\\
\hline
HF & 70& 46& 0.032 & 142& 121& 0.034 & 71& 66& 0.032\\
\hline
PI & 151& 89& 0.037 & 301& 190& 0.036 & 88& 67& 0.035\\
\hline
COVID 6hr & 1001& 401& 0.031 & 2325& 998& 0.033 & 142& 71& 0.031\\
\hline
COVID 12hr & 1001& 401& 0.031 & 2325& 998& 0.033 & 142& 71& 0.031\\
\hline
COVID 18hr & 1001& 401& 0.031 & 2325& 998& 0.033 & 142& 71& 0.031\\
\hline
COVID 24hr & 1001& 401& 0.031 & 2325& 998& 0.033 & 142& 71& 0.031\\
\hline
COVID 30hr & 1001& 401& 0.031 & 2325& 998& 0.033 & 142& 71& 0.031\\
\hline
COVID 36hr & 1001& 401& 0.031 & 2325& 998& 0.033 & 142& 71& 0.031\\
\hline
COVID 42hr & 1001& 401& 0.031 & 2325& 998& 0.033 & 142& 71& 0.031\\
\hline
COVID 48hr & 1001& 401& 0.031 & 2325& 998& 0.033 & 142& 71& 0.031\\
\hline
COVID 54hr & 1001& 401& 0.031 & 2325& 998& 0.033 & 142& 71& 0.031\\
\hline
COVID 60hr & 1001& 401& 0.031 & 2325& 998& 0.033 & 142& 71& 0.031\\
\hline
COVID 66hr & 1001& 401& 0.031 & 2325& 998& 0.033 & 142& 71& 0.031\\
\hline
COVID 72hr & 1001& 401& 0.031 & 2325& 998& 0.033 & 142& 71& 0.031\\
\hline
Diabetes & 68& 35& 0.034 & 121& 101& 0.043 & 70& 69& 0.035\\
\hline
CAD & 98& 79& 0.032 & 132& 97& 0.031 & 73& 66& 0.038\\
\hline
\end{tabular}
\end{adjustbox}
\end{table*}

\section{Conclusion}

Healthcare data is complex and have many concurrences and loops. This makes predicting healthcare outcomes difficult, because data quality is poor. By using concatenation algorithm as a pre-processing step, data complexity decreased, improving the performance of process discovery algorithms by recommending a probability-based concatenation of events, with concurrent relations. Thus, it can predict critical healthcare outcomes more accurately.

Predicting critical health outcomes allows medical stakeholders to provide optimal treatment for patients to increase life expectancy. For instance, if a patient is predicted to die or readmit, investing many resources on them will not be optimal. Moreover, in cases when we can predict a patient's outcome sooner, medical stakeholders can expedite treatment to save a patient's life. Furthermore, knowing outcomes with high AUC would allow medical teams improve treatment decisions and allocate medical resources more effectively. 

We used 16 healthcare datasets and fed them to the SM algorithm to discover the process model. Moreover, the pre-processing step was also applied to the same event logs, and the resultant event logs were used by the SM algorithm to discover a process model. We then measured the F-Measure and complexity of the models for both cases. We fed both pre-processed and raw data to the DREAM algorithm for prediction, and compared the results. In cases where improvements were observed in evaluation metrics, the Wilcoxon Test was run to measure the level of significance. Results indicated that the pre-processing step increased F-Measure values and decreased the complexity of the models statistically. Prediction of critical health outcomes improved as well. These results demonstrate that using a pre-processing step improves data quality, hence improves the process model and prediction modeling results. Using an effective process mining / deep learning algorithm like the DREAM algorithm also, ensures medical history and time information related to variables are not ignored which can predict outcomes more accurately.

In future work, we plan to use other pre-processing approaches to see their effects on data quality and predicting critical health outcomes. We plan to find techniques that optimally adjust parameter values automatically, rather than using hyper-parameter optimization currently in favor. We would also validate the proposed approach using healthcare datasets. Finally, MIMIC-III and UIH datasets contains useful information like clinical notes and images that can be converted into event logs and fed to models as inputs.
\appendix

\section*{Appendix: Notations}
\begin{description}[labelwidth=1.5cm, labelsep=0.3cm, leftmargin=2cm, style=multiline]
    \item[$a$] Event
    \item[$A$] Finite set of all events
    \item[$\alpha$] Decay rate
    \item[$\alpha_p$] Decay rate for a specfic place p
    \item[$\beta$] Constant parameter of a decay function
    \item[$C(\tau)$] Token counting vector from time 0 to $\tau$ each element represents the number of tokensthat entered a specific place
    \item[$d$] Attribute
    \item[$d_{ts}$] Timestamp attribute
    \item[$D$] Finite set of all possible attributes
    \item[$\Delta_{max}(L)$] Maximum observed trace duration in an event log $L$
    \item[$\Delta_p$] Difference of $\tau$ and $\tau_p$
    \item[$\delta_p(g)$] function of average time between a token is consumed in place p until a new token is produced
in p based on an input trace g
    \item[$E$] Event instance vector
    \item[$e$] An instance of E
    \item[$\eta$] SM threshold which controls concurrency relations
    \item[$\epsilon$] SM threshold which controls filtering process
    \item[$f_p(\tau)$] Decay function of place $p$
    \item[$F(\tau)$] Decay function response vector
    \item[$g$] Case or trace
    \item[$G$] Finite set of all possible traces
    \item[$L$] Event log which is a set of traces
    \item[$L_{i,j}$] $j^\textit{th}$ event instance in the $i^\textit{th}$ trace of an event log $L$
    \item[$\ell$] label of each event
    \item[$\lambda$] function that retrieves the label of each event
    \item[$\lambda_p(g)$] Number of tokens a place $p$ produces when replaying a trace $g$
    \item[$M(\tau)$] Vector representing the marking of a PN
    \item[${mean()}$] \hspace{-0.1cm} Arithmetic mean function
    \item[$N$] Set of all possible event instances
    \item[$\nu_d(E)$] Function returning value of attribute $d$ of event instance $E$
    \item[$p$] Place
    \item[$P$] Set of all places
    \item[$PN$] Mathematical definition of a labeled PN
    \item[$p\star$] Threshold value for concatenation algorithm
    \item[$S(\tau)$] Timed state sample at time $\tau$
    \item[$S$] Set of timed state samples
    \item[$\tau$] Time
    \item[$\tau_p$] Most recent time that a token entered place $p$
    \item[$v_p(g)$] number of tokens a place p produces when replaying a trace g
    \item[$e_i \oplus e_j$] Concatenated event for $e_i$ and $e_j$
    \item[$p(e_i \rightarrow e_j)$]The probability that ej happens instantly after ei
    \item[$p(e_i||e_j)$] probability of the frequency that event ej which occurs immediately after event ei
\end{description}

\section*{Acknowledgment}
The authors would like to thank all contributors of MIMIC-III for providing the public EHR dataset.

\end{document}